\documentclass[conference,10pt]{IEEEtran}

\usepackage[T1]{fontenc}
\usepackage{times}
\usepackage[letterpaper, left=0.625in, right=0.625in, top=0.75in, bottom=1.0in]{geometry}

\usepackage{booktabs}
\usepackage{stix}

\usepackage{amsfonts}
\usepackage{xcolor,soul,framed} 

\colorlet{shadecolor}{yellow}
\usepackage[pdftex]{graphicx}
\graphicspath{{../pdf/}{../jpeg/}}
\DeclareGraphicsExtensions{.pdf,.jpeg,.png}

\usepackage[cmex10]{amsmath}
\usepackage[compatibility=false]{caption}
\usepackage[T1]{fontenc}
\usepackage{multirow}

\usepackage{amsmath}

\usepackage{array}
\usepackage{mdwmath}
\usepackage{mdwtab}
\usepackage{eqparbox}
\usepackage{url}
\usepackage{blkarray}

\usepackage{amsfonts}
\usepackage{mathtools}

\usepackage{algorithm}
\usepackage{algpseudocode}
\usepackage{float} 
\usepackage{subcaption}

\usepackage[T1]{fontenc}  

\usepackage{etoolbox}
\makeatletter
\patchcmd{\ALG@doentity}{\noindent}{}{}{}
\makeatother
\usepackage{mathrsfs}

\usepackage[colorlinks=true, linkcolor=black, citecolor=black, urlcolor=black, bookmarks=false]{hyperref}

\algnewcommand\Output{\textbf{Output: }}

\IEEEoverridecommandlockouts
 
\begin{document}
 \setlength{\columnsep}{0.21in}
     \title{Adaptive Role Assignment and Controlled Feature Reconstruction for Federated Image Semantic Communication}
%





\author{
\IEEEauthorblockN{
Yuxuan Chen$^{\dagger}$, 
Jiao Bie$^{\dagger}$,
Hang Qi,
Jingjin Wu\\
}

\IEEEauthorblockA{
Guangdong Prov./Zhuhai Key Lab of IRADS, 
Beijing Normal-Hong Kong Baptist University, Guangdong, P. R. China \\
}

Email: t330018009@mail.bnbu.edu.cn;
t330018001@mail.bnbu.edu.cn;\\
t330202706@mail.bnbu.edu.cn;
jj.wu@ieee.org\\

\thanks{$^{\dagger}$These authors contributed equally to this work. This work was partly supported 
the Guangdong Provincial Key Laboratory IRADS, BNBU (Project code 2022B1212010006), 
and Guangdong Higher Education Upgrading Plan (2021-2025) UIC R0400001-22 and R0400024-22.}
}

\maketitle

%
\begin{abstract}
Hybrid federated semantic communication enables user equipment (UEs) to contribute model updates or semantic features under a shared training framework.
However, heterogeneous role costs and the interaction between model aggregation and feature reconstruction (FR) remain key challenges in practical radio access network (RAN) deployment.
To address these issues, we propose ARC-FedSC, a RAN-aware framework that coordinates UE role assignment and server-side controlled FR under a fixed communication budget. Within ARC-FedSC, RAN Aware Role-Bandit learns the historical contribution of UEs in different roles together with current uplink conditions, while Temporal Consistency Alignment (TCA) regulates FR according to recent model aggregation directions. Experiments on CIFAR-10 with time-varying RAN channels show that ARC-FedSC outperforms baselines across all five tested role configurations in the later training stage, with the best configurations achieving about $24.17$ dB average PSNR over the last 20 rounds. Component level results further show that Role-Bandit maintains an average rolling UE coverage of $85.3\%$, compared with about $65.7\%$ for baseline, while TCA reduces the probability of harmful FR events from $0.245$ to $0.093$, corresponding to a $62.0\%$ reduction. Ablation and sensitivity studies further support the effectiveness and robustness of the proposed design.

\end{abstract}


%
\IEEEpeerreviewmaketitle


\section{Introduction}

In future RANs, semantic communication has emerged as a promising paradigm for improving communication efficiency by focusing on the delivery of information relevant to the communication objective~\cite{xie2021deep}. 
For image transmission, deep joint source and channel coding (JSCC) learns the encoder and decoder in an end to end manner, enabling source images to be directly mapped into channel symbols and reconstructed at the receiver~\cite{bourtsoulatze2019deep,dai2022nonlinear}. Since the performance of deep JSCC relies on learned neural models, these models may require
periodic updates after deployment to maintain their performance in practical wireless networks~\cite{huh2025feature}.
Federated learning (FL) provides a distributed approach for collaboratively updating shared models while keeping raw user data at local devices~\cite{mcmahan2017communication,Qi2025}, and has recently been introduced into semantic communication systems to enable collaborative model training across distributed UEs ~\cite{sun2024federated,nguyen2024efficient,yan2024fssc}. Recent studies have further explored federated semantic communication in dynamic applications such as UAV swarms and vehicular networks~\cite{xu2024federated,zheng2024mobility,yu2026scusfl}. In practical cellular edge networks with mobile cameras, varying uplink conditions may prevent some UEs from transmitting model updates within the required deadline, highlighting the importance of communication-efficient collaborative training.

However, applying FL to wireless semantic communication incurs considerable uplink communication overhead because participating UEs repeatedly transmit model updates over limited radio resources~\cite{amiri2020federated}. 
Accordingly, client selection and radio resource allocation have been widely studied to improve the efficiency of wireless FL~\cite{nishio2019client,xu2020client}. Recent works further couple client selection with communication efficient model transmission. For example, FedCG jointly considers client selection and gradient compression under heterogeneous edge conditions~\cite{xu2023federated}, while MFedMC jointly selects clients and modality encoders by considering learning contribution, communication cost, and historical recency~\cite{yuan2026communication}. In semantic communication, hybrid learning frameworks have explored the joint use of model updates and semantic information for collaborative training~\cite{sun2025hybrid}, while heterogeneous IoT terminals have been coordinated for model training and semantic extraction~\cite{xu2026flsc}. For image semantic communication, FedSFD employs semantic feature distillation~\cite{10599557}, while FedSFR further adopts a hybrid architecture in which model role UEs transmit sparsified local updates and feature role UEs transmit compact semantic features to the server~\cite{huh2025feature}. The server aggregates the received model updates and subsequently performs FR using the received features.

Despite these advances, two issues remain insufficiently addressed in hybrid federated semantic communication. 
First, existing studies have investigated utility based client selection, communication efficient model transmission, and hybrid semantic learning~\cite{yuan2026communication,lai2021oort,xia2020multi}.
However, they do not explicitly consider role-specific historical contributions for model and feature transmission under a shared RAN communication budget. Although FedSFR supports both roles, its role assignment mainly depends on current channel conditions and does not learn the historical contribution of individual UEs in different roles~\cite{huh2025feature}.
Thus, how to jointly exploit role-specific historical contributions and current uplink radio conditions under a common communication budget remains insufficiently explored. Second, although FedSFR analyzes the directional relation between the server-side update and sparsification error~\cite{huh2025feature}, it does not explicitly coordinate FR with the preceding model aggregation during training. Consequently, the realized server-side FR update may partially counteract the aggregation update. These limitations raise a central question: how can heterogeneous UE contributions from different transmission roles be effectively selected under constrained RAN resources and consistently integrated during federated semantic model training?

\begin{figure*}[t]
    \centering
    \includegraphics[width=0.98\textwidth]{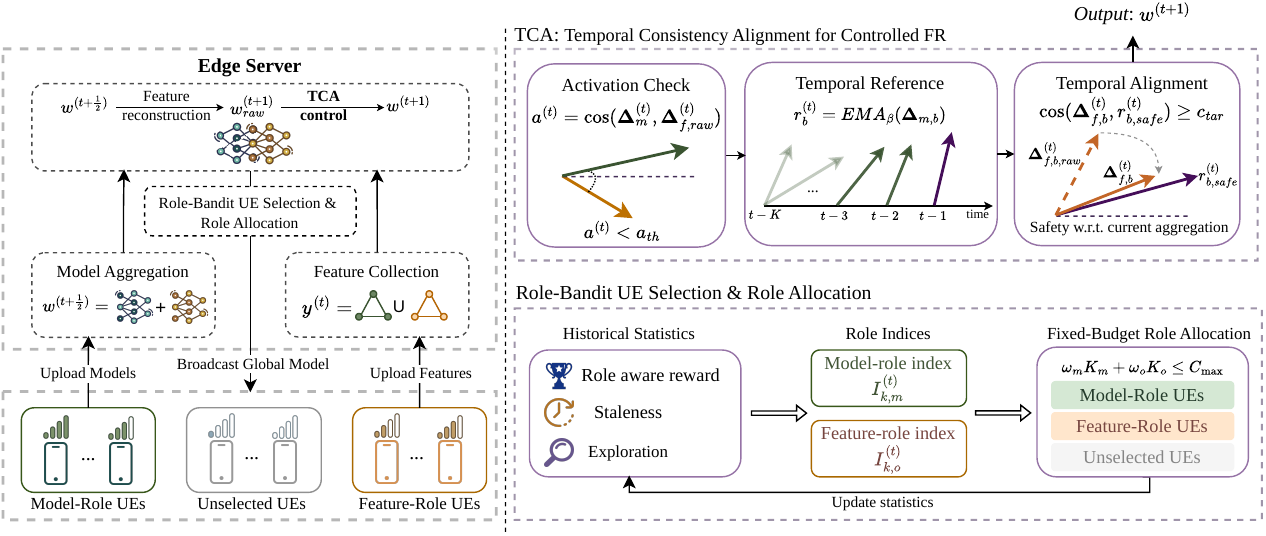}
    \caption{Overall architecture of the proposed ARC-FedSC framework, including the hybrid FL and FR architecture, Temporal Consistency Alignment, and Role-Bandit based UE selection and role allocation.}
    \label{fig:framework}
\end{figure*}

To address these issues, we propose ARC-FedSC, which
coordinates UE participation and server-side FR within a unified federated semantic learning process. ARC-FedSC consists of two components.
For UE participation, ARC-FedSC maintains separate historical utility estimates for the model and feature roles, and jointly determines UE role assignments according to their learned contributions, instantaneous RAN feasibility, and the shared communication budget. For server-side updating, ARC-FedSC uses recent model aggregation directions to construct a temporal reference for FR and corrects the raw FR direction when persistent inconsistency is detected.
The overall architecture of ARC-FedSC is illustrated in Fig.~\ref{fig:framework}.

The main contributions of this work are summarized as follows:
\begin{itemize}
    \item We formulate UE selection and role assignment for hybrid federated semantic communication under a common RAN communication budget, where model and feature roles have different transmission costs and deadline constraints. We then develop a RAN Aware Role-Bandit mechanism that learns the historical contribution of each UE in different roles and combines it with current uplink radio conditions for role assignment.

    \item We develop TCA for controlled FR after model aggregation. By incorporating temporal information from
    previous aggregation updates, TCA reduces the conflict between model aggregation and subsequent FR, thereby improving the
    reliability of server-side FR.
    
    \item We conduct extensive experiments under time-varying RAN channels with multiple role configurations and random seeds. Under the same communication budget, ARC-FedSC achieves better reconstruction performance than baselines in the later training stage and requires fewer training rounds to reach the target PSNR. It also enables broader UE participation and reduces harmful FR events. Ablation and sensitivity studies further demonstrate the contributions of both modules in ARC-FedSC and their robustness to the tested parameter settings.
\end{itemize}


\section{System Model}
\label{model}

\subsection{Federated Semantic FR}

We consider a parameter server and $K$ clients
$\mathcal{A}=\{1,\ldots,K\}$. Client $k$ owns a local dataset
$\mathcal{D}_k$, which contains a public subset
$\mathcal{P}_k\subseteq\mathcal{D}_k$ used for semantic feature
upload. The global semantic model is parameterized by
$\mathbf{w}$ and consists of the JSCC encoder and decoder. Its local
training objective at client $k$ is
\[
F_k(\mathbf{w})
=
\frac{1}{|\mathcal{D}_k|}
\sum_{\mathbf{X}\in\mathcal{D}_k}
\ell_c(\mathbf{w};\mathbf{X}),
\]
where $\ell_c(\cdot)$ denotes the end-to-end reconstruction loss.
The global objective is
\begin{equation}
F(\mathbf{w})
=
\sum_{k\in\mathcal{A}}p_kF_k(\mathbf{w}),
\qquad
p_k
=
\frac{|\mathcal{D}_k|}
{\sum_{j\in\mathcal{A}}|\mathcal{D}_j|}.
\end{equation}

At global round $t$, the server broadcasts the current model
$\mathbf{w}^{(t)}$ to the selected clients. Each selected client
performs $E_c$ local epochs on its own dataset and obtains the locally
updated model $\mathbf{w}_k^{(t,E_c)}$. According to its assigned role,
the client then uploads either a sparse model update or semantic
features.

For a model-role client $k_m\in\mathcal{A}_m^{(t)}$, the local model
difference is
$\boldsymbol{\Delta}_k^{(t)}
=\mathbf{w}^{(t)}-\mathbf{w}_k^{(t,E_c)}$.
To reduce uplink communication overhead, the update is compressed
using Top-$S$ sparsification with error feedback~\cite{amiri2020federated}.
Specifically, the accumulated compression residual is added before
sparsification, while the discarded components are retained for the
next round. Let $\mathbf{g}_k^{(t)}$ denote the resulting sparse
update and $M_t=|\mathcal{A}_m^{(t)}|$. The model obtained after aggregating the sparse updates from
model-role clients is denoted by
$\mathbf{w}^{(t+\frac{1}{2})}$, representing the intermediate
server state between global rounds $t$ and $t+1$ prior to feature
reconstruction. Accordingly, the server first aggregates
the received model updates as
\begin{equation}
\mathbf{w}^{(t+\frac{1}{2})}
=
\mathbf{w}^{(t)}
-
\frac{K}{M_t}
\sum_{k\in\mathcal{A}_m^{(t)}}
p_k\mathbf{g}_k^{(t)}.
\end{equation}

For a feature-role client $k_o\in\mathcal{A}_o^{(t)}$, the locally
updated encoder is applied to the public subset $\mathcal{P}_k$ to
generate semantic feature vectors. Instead of uploading model
parameters, the client transmits only these features to the server.
Let $\mathcal{D}_s^{(t)}$ denote the collection of semantic features
received from all feature-role clients in round $t$. For each received
feature vector $\mathbf{y}$, the server reconstructs it through the
JSCC model as
\begin{equation}  
\widehat{\mathbf{y}}
=
f_{\boldsymbol{\theta}}
\bigl(
f_{\boldsymbol{\phi}}^{-1}
(\widetilde{\mathbf{y}}+\mathbf{n})
\bigr), \quad \widetilde{\mathbf{y}} = \frac{\mathbf{y}}{\|\mathbf{y}\|_2}
\end{equation}
where $\boldsymbol{\theta}$ and $\boldsymbol{\phi}$ denote the corresponding encoder and decoder parameters, respectively, and $\mathbf{n} \sim \mathcal{N}(\mathbf{0}_d, \sigma^2 \mathbf{I}_d)$ denotes
channel noise. Starting from the aggregated model
$\mathbf{w}^{(t+\frac{1}{2})}$, the server performs feature
reconstruction by minimizing
\begin{equation}
F_s^{(t)}(\mathbf{w}^{(t+\frac{1}{2})})
=
\frac{1}{|\mathcal{D}_s^{(t)}|}
\sum_{\mathbf{y}\in\mathcal{D}_s^{(t)}}
\operatorname{MSE}
\left(
\widehat{\mathbf{y}},\mathbf{y}
\right).
\end{equation}

After $E_s$ server-side reconstruction epochs, the resulting model is
denoted by $\mathbf{w}_{\mathrm{raw}}^{(t+1)}$. Hence, each global
round consists of two successive server updates,
\[
\mathbf{w}^{(t)}
\longrightarrow
\mathbf{w}^{(t+\frac{1}{2})}
\longrightarrow
\mathbf{w}_{\mathrm{raw}}^{(t+1)},
\]
where the first stage aggregates sparse model updates and the second
stage incorporates the semantic information uploaded by feature-role
clients. The subsequent TCA stage determines whether the raw FR update should be directly accepted or directionally adjusted into $\mathbf{w}^{(t+1)}$, as detailed in Section~\ref{sec:tca}. This sequential update motivates controlled FR, since the raw FR step may not always be consistent with the preceding model aggregation update.

\subsection{RAN Uplink and Fixed Role Budget}

We consider an orthogonal RAN uplink, where each scheduled UE is
allocated a dedicated bandwidth $B$ and transmits in parallel without
inter-UE interference. To capture temporal channel variation, the
uplink channel of UE $k$ follows a first-order Gauss--Markov model:
\begin{equation}
h_k^{(t)}
=
\rho h_k^{(t-1)}
+
\sqrt{1-\rho^2}\,\epsilon_k^{(t)},
\qquad
\epsilon_k^{(t)}\sim\mathcal{CN}(0,1),
\end{equation}
where $\rho\in[0,1]$ controls the temporal channel correlation. Given
the average SNR $\bar{\gamma}_k$, the instantaneous uplink SNR and
achievable rate are
\begin{equation}
\gamma_k^{(t)}
=
\bar{\gamma}_k|h_k^{(t)}|^2,
\qquad
R_k^{(t)}
=
B\log_2\left(1+\gamma_k^{(t)}\right).
\end{equation}

In each round, a selected UE can upload either a sparse model update
or semantic feature vectors, corresponding to the model role
$r=m$ and feature role $r=o$, respectively. Let $L_m$ and $L_o$
denote their payload sizes. The transmission latency of UE $k$
when assigned to role $r\in\{m,o\}$ is
\begin{equation}
T_{k,r}^{(t)}
=
\frac{L_r}{R_k^{(t)}}.
\end{equation}
A UE-role pair $(k,r)$ is feasible only if
$T_{k,r}^{(t)}\leq T_{\max}$, where $T_{\max}$ is the hard uplink
deadline. Hence, the feasible roles of a UE may vary across rounds
with its instantaneous channel condition.

Let $\mathcal{A}_m^{(t)}$ and $\mathcal{A}_o^{(t)}$ denote the
model-role and feature-role UE sets at round $t$. Each UE can be
assigned to at most one role, and the selected sets satisfy
\begin{equation}
\label{equ:selected_sets}
|\mathcal{A}_m^{(t)}|\leq K_m,
\qquad
|\mathcal{A}_o^{(t)}|\leq K_o,
\qquad
\mathcal{A}_m^{(t)}\cap\mathcal{A}_o^{(t)}=\varnothing.
\end{equation}

Since sparse model updates and semantic features incur different uplink communication costs, we associate the model and feature roles with normalized costs $\omega_m$ and $\omega_o$, respectively. The fixed role limits $(K_m,K_o)$ therefore satisfy
\begin{equation}
\label{equ:budget}
\omega_m K_m+\omega_o K_o\leq C_{\max}.
\end{equation}

The normalized costs characterize the relative communication demands of the two upload modes under the fixed transmission settings considered in this work. While $(K_m,K_o)$ remains fixed for a given configuration, the specific UEs and their assigned roles are determined in each round according to the instantaneous RAN state and their learning values. Consequently, the actual numbers of selected UEs may be smaller than the prescribed role limits when insufficient UE-role pairs satisfy the transmission deadline.

\subsection{Problem Formulation}

Let $G$ denote the total number of global training rounds. At global round $t$,
$z_{k,m}^{(t)},z_{k,o}^{(t)}\in\{0,1\}$ indicate whether
UE $k$ is assigned to the model role or the feature role,
respectively, and define $\mathcal{A}_r^{(t)} = \{k\in\mathcal{A}:z_{k,r}^{(t)}=1\}, \,r\in\{m,o\}$.
The objective is to minimize the expected global image reconstruction loss at the end of the training horizon:
\begin{equation}
\begin{aligned}
    \min_{\{z_{k,m}^{(t)},z_{k,o}^{(t)}\}_{t=0}^{G-1}}
    \quad&
    \mathbb{E}
    \left[
        F\left(\mathbf{w}^{(G)}\right)
    \right] \\
    \mathrm{s.t.}\quad
    &
    z_{k,r}^{(t)}T_{k,r}^{(t)}
    \leq T_{\max},
    &&\forall k,t,\ r\in\{m,o\},\\
\end{aligned}
\end{equation}
The induced role sets $\mathcal{A}_m^{(t)}$ and $\mathcal{A}_o^{(t)}$ further satisfy the UE exclusivity,
role-limit, and communication-budget constraints in \eqref{equ:selected_sets} and \eqref{equ:budget}.

The expectation is taken over the time-varying RAN channels
and the stochastic training process. Since long-term contribution of UE role assignment to the final reconstruction loss cannot be directly observed during online training, we use the role-specific leave-one-out (LOO) PSNR gain as a tractable surrogate for its marginal learning contribution. The historical LOO gains are then incorporated into the Role-Bandit index to guide online UE selection and role assignment, as detailed in Section~\ref{sec:role_bandit}.

\section{Algorithms}
\subsection{RAN Aware Role-Bandit UE Selection and Role Assignment}
\label{sec:role_bandit}

The same UE may provide different learning contributions when assigned to the model role or the feature role. Therefore, we maintain independent role-specific statistics for each UE and estimate its contribution from historical training feedback.

For a selected UE $k$, its role-specific contribution is evaluated by the LOO PSNR gain as
\begin{equation}
R_{k,m}^{(t)}=P_{\mathrm{agg}}^{(t)}-P_{\mathrm{agg},-k}^{(t)},
\qquad
R_{k,o}^{(t)}=P_{\mathrm{FR}}^{(t)}-P_{\mathrm{FR},-k}^{(t)},
\end{equation}
where $P_{\mathrm{agg}}^{(t)}$ and $P_{\mathrm{FR}}^{(t)}$ denote the PSNR after model aggregation and raw FR, respectively, and the subscript $-k$ denotes the corresponding result without UE $k$. 

To control the computational overhead of role-specific reward estimation, LOO rewards are updated only during periodic calibration rounds. For model-role UEs, the server reuses the received sparse updates to construct LOO aggregates and evaluates their PSNR on the server anchor set. In contrast, evaluating the LOO reward of a feature-role UE requires an additional server-side FR pass after excluding its uploaded features, which incurs a higher computational cost. Therefore, all selected model-role UEs are evaluated during calibration, whereas only a capped subset of selected feature-role UEs is evaluated in each calibration round, prioritizing UE-role pairs with fewer previous reward observations. The FR result obtained using the complete feature set is retained for the subsequent TCA stage. After the initial coverage stage, calibration is performed every five global rounds, amortizing the additional server-side computation over the training process.

For role $r\in\{m,o\}$, the LOO reward $R_{k,r}^{(t)}$ is clipped to
$[-R_{\max},R_{\max}]$, yielding
$\widetilde{R}_{k,r}^{(t)}$.
Let $Q_{k,r}^{(t)}$ and $N_{k,r}^{(t)}$ denote the discounted reward
sum and observation count, respectively:
\begin{equation}
\begin{aligned}
Q_{k,r}^{(t)}
&=\lambda Q_{k,r}^{(t-1)}
+\delta_{k,r}^{(t)}\widetilde{R}_{k,r}^{(t)},\\
N_{k,r}^{(t)}
&=\lambda N_{k,r}^{(t-1)}
+\delta_{k,r}^{(t)},
\end{aligned}
\end{equation}
where $\lambda\in(0,1)$ is the discount factor and $\delta_{k,r}^{(t)}$ indicates whether the reward is observed at round $t$. The discounted mean reward is defined as $\widehat{\mu}_{k,r}^{(t)}=Q_{k,r}^{(t)}/N_{k,r}^{(t)}$ with $\widehat{\mu}_{k,r}^{(t)}=0$ when $N_{k,r}^{(t)}=0$. We construct the Role-Bandit index as
\begin{equation}
I_{k,r}^{(t)}
=
\widehat{\mu}_{k,r}^{(t)}
+
U_{k,r}^{(t)}
+
S_{k,r}^{(t)}
-
C_{k,r}^{(t)},
\end{equation}
where $U_{k,r}^{(t)}$ is the exploration term, $S_{k,r}^{(t)}$ is the role-specific staleness bonus, and $C_{k,r}^{(t)}$ is the communication penalty.

Inspired by the UCB principle~\cite{auer2002finite}, we define the role-specific exploration term as
\begin{equation}
U_{k,r}^{(t)}
=
\beta_r
\sqrt{
\frac{
\log\left(N_r^{(t)}+1\right)
}{
N_{k,r}^{(t)}+\nu
}
},
\end{equation}
where $N_r^{(t)}=\sum_{k\in\mathcal{A}}N_{k,r}^{(t)}$
denotes the total discounted observation count for role $r$, and $\beta_r$ controls the exploration strength.
The constant $\nu>0$ prevents an excessively small denominator for rarely observed UE role pairs. The staleness and communication terms are given by
\begin{equation}
S_{k,r}^{(t)}
=
s_r
\min
\left\{
\frac{\Delta_{k,r}^{(t)}}{H},1
\right\},
\qquad
C_{k,r}^{(t)}
=
c_r
\min
\left\{
\frac{T_{k,r}^{(t)}}{T_{\max}},1
\right\},
\end{equation}
where $\Delta_{k,r}^{(t)}$ denotes the role-specific staleness, $H$ is its saturation horizon, and $s_r$ controls the maximum staleness bonus. Moreover, $c_r$ is the role-specific communication penalty coefficient, while $T_{k,r}^{(t)}/T_{\max}$ measures the normalized uplink latency. Hence, UEs with longer transmission latency receive a larger communication penalty. A UE role pair violating the hard deadline is regarded as infeasible.


Based on the resulting role indices, the model- and feature-role UE sets are jointly determined by dynamic programming. Let $D_k(m,o,c)$ denote the maximum accumulated Role-Bandit index after considering the first $k$ UEs, where $m$ and $o$ are the numbers of model- and feature-role UEs, respectively, and $c$ is the consumed communication budget. For each UE, three actions are considered: not selected, assigned to the model role, or assigned to the feature role. The state transition is
\begin{equation}
\begin{aligned}
D_k(m,o,c)
={}&
\max_{r\in\{\varnothing,m,o\}}
\{
D_{k-1}\bigl(
m-\mathbf{1}(r=m),\,
o-\mathbf{1}(r=o),\\
&\qquad\qquad
c-\omega_r
\bigr)
+ I_{k,r}^{(t)}
\} .
\end{aligned}
\label{eq:dp}
\end{equation}
where $\omega_{\varnothing}=0$ and $I_{k,\varnothing}^{(t)}=0$. Transitions assigning UE $k$ to role $r$ are allowed only when the corresponding UE-role pair satisfies the transmission deadline.

\begin{algorithm}[t]
\caption{RAN Aware Role-Bandit UE Selection and Role Assignment}
\label{alg:role_bandit}
\begin{algorithmic}[1]
\Require UE set $\mathcal{A}$, total global rounds $G$, role limits $(K_m,K_o)$, role costs $(\omega_m,\omega_o)$, communication budget $C_{\max}$, deadline $T_{\max}$

\State Initialize $Q_{k,r}$, $N_{k,r}$, and role-specific staleness
for all $k\in\mathcal{A}$ and $r\in\{m,o\}$

\For{$t=0,\ldots,G-1$}

    \State Obtain the instantaneous RAN state and compute
    $T_{k,r}^{(t)}$ for each UE-role pair
    \State Exclude $(k,r)$ if $T_{k,r}^{(t)}>T_{\max}$

    \State Discount $Q_{k,r}^{(t)}$ and $N_{k,r}^{(t)}$
    using factor $\lambda$

    \If{$t$ is in the initial coverage stage}
        \State Select feasible UE-role pairs to improve
        model- and feature-role coverage under the role and budget constraints
    \Else
        \ForAll{feasible $(k,r)$}
            \State Compute
            $\widehat{\mu}_{k,r}^{(t)},\, U_{k,r}^{(t)}, \,S_{k,r}^{(t)} and C_{k,r}^{(t)}$
            
            \State Form
            $I_{k,r}^{(t)}
            =\widehat{\mu}_{k,r}^{(t)}
            +U_{k,r}^{(t)}
            +S_{k,r}^{(t)}
            -C_{k,r}^{(t)}$
        \EndFor

        \State Use dynamic programming to assign each UE to
        \emph{model role}, \emph{feature role}, or \emph{not selected},
        subject to UE exclusivity and
        $|\mathcal{A}_m^{(t)}|\leq K_m$,
        $|\mathcal{A}_o^{(t)}|\leq K_o$

        \State Among feasible assignments, prioritize
        $\omega_m |\mathcal{A}_m^{(t)}|+\omega_o |\mathcal{A}_o^{(t)}|$,
        then $|\mathcal{A}_m^{(t)}|+|\mathcal{A}_o^{(t)}|$,
        and finally
        $\sum_{k,r} I_{k,r}^{(t)}$

        \State Recover the selected sets
        $\mathcal{A}_m^{(t)}$ and $\mathcal{A}_o^{(t)}$
    \EndIf

    \State Perform model aggregation using $\mathcal{A}_m^{(t)}$
    and raw FR using $\mathcal{A}_o^{(t)}$

    \If{$t$ is a calibration round}
        \State Compute the LOO rewards
        $R_{k,m}^{(t)}$ and $R_{k,o}^{(t)}$
        for the corresponding selected UEs
        \State Clip the observed rewards and update
        $Q_{k,r}^{(t)}$ and $N_{k,r}^{(t)}$
    \EndIf

    \State Reset the staleness of selected UE-role pairs and
    increase it for the remaining pairs

\EndFor

\State \Return
$\{\mathcal{A}_m^{(t)},\mathcal{A}_o^{(t)}\}_{t=0}^{G-1}$
\end{algorithmic}
\end{algorithm}

\subsection{TCA Controlled Feature Reconstruction}
\label{sec:tca}

Although model aggregation and FR both aim to
improve the global semantic model, they incorporate knowledge from
different UEs and learning processes. As a result, the raw FR update
may deviate from the model aggregation direction and degrade the
knowledge obtained from model-role UEs. Moreover, the aggregation
direction itself may fluctuate across rounds due to dynamic UE
participation and stochastic local training. To stabilize the FR
process, we propose Temporal Consistency Alignment (TCA), which
controls the FR direction according to the recent model aggregation
trend.

We first define the model aggregation direction and the raw FR
direction as
\begin{equation}
\label{equ:tca_direction}
\boldsymbol{\Delta}_m^{(t)}
=
\mathbf{w}^{(t)}
-
\mathbf{w}^{(t+\frac{1}{2})},
\qquad
\boldsymbol{\Delta}_{f,\mathrm{raw}}^{(t)}
=
\mathbf{w}^{(t+\frac{1}{2})}
-
\mathbf{w}_{\mathrm{raw}}^{(t+1)}.
\end{equation}
Their directional consistency is measured by
\begin{equation}
\label{equ:tca_cos}
a^{(t)}
=
\cos\left(
\boldsymbol{\Delta}_m^{(t)},
\boldsymbol{\Delta}_{f,\mathrm{raw}}^{(t)}
\right).
\end{equation}
To avoid unnecessary intervention, TCA is activated only when
$a^{(t)}<a_{\mathrm{th}}$ persists for $P_{\mathrm{act}}$
consecutive rounds, where $a_{\mathrm{th}}$ and $P_{\mathrm{act}}$
denote the activation threshold and patience, respectively. Before
activation, the raw FR update is directly accepted. Once activated,
TCA remains active in subsequent rounds.

Since a single-round aggregation direction may be unstable, TCA
maintains an exponential moving average (EMA) of recent aggregation
directions for each model block $b\in\{\mathrm{enc},\mathrm{dec}\}$:
\begin{equation}
\label{equ:tca_ema}
\mathbf{h}_b^{(t)}
=
\beta\mathbf{h}_b^{(t-1)}
+
(1-\beta)
\frac{
\boldsymbol{\Delta}_{m,b}^{(t)}
}{
\left\|\boldsymbol{\Delta}_{m,b}^{(t)}\right\|_2},
\qquad
\mathbf{r}_b^{(t)}
=
\frac{
\mathbf{h}_b^{(t)}
}{
\left\|\mathbf{h}_b^{(t)}\right\|_2},
\end{equation}
where $\beta\in[0,1)$ controls the contribution of historical directions, and $\mathbf{r}_b^{(t)}$ serves as the temporal reference.

To prevent an outdated temporal reference from conflicting with the current aggregation direction, we construct the safe reference
\begin{equation}
\label{equ:tca_safe_ref}
\mathbf{r}_{b,\mathrm{safe}}^{(t)}
=
\begin{cases}
\mathbf{r}_b^{(t)},
&
\cos\left(
\boldsymbol{\Delta}_{m,b}^{(t)},
\mathbf{r}_b^{(t)}
\right)\geq 0,
\\[2mm]
\dfrac{
\boldsymbol{\Delta}_{m,b}^{(t)}
}{
\|\boldsymbol{\Delta}_{m,b}^{(t)}\|_2
},
&
\text{otherwise},
\end{cases}
\end{equation}
where the second case is also used when the temporal reference is unavailable.

For each block, TCA preserves the raw FR direction if its cosine
similarity with the safe reference is at least $c_{\mathrm{tar}}$.
Otherwise, the raw FR direction is adjusted toward
$\mathbf{r}_{b,\mathrm{safe}}^{(t)}$ while preserving its norm until
the target cosine similarity $c_{\mathrm{tar}}$ is reached. We denote
the resulting temporally aligned FR direction by
$\overline{\boldsymbol{\Delta}}_{f,b}^{(t)}$.

Finally, to prevent temporal alignment from introducing a conflict
with the current aggregation direction, we apply
\begin{equation}
\label{equ:tca_safety}
\widetilde{\boldsymbol{\Delta}}_{f,b}^{(t)}
=
\overline{\boldsymbol{\Delta}}_{f,b}^{(t)}
-
\frac{
\min\left\{
\left\langle
\overline{\boldsymbol{\Delta}}_{f,b}^{(t)},
\boldsymbol{\Delta}_{m,b}^{(t)}
\right\rangle,
0
\right\}
}{
\|\boldsymbol{\Delta}_{m,b}^{(t)}\|_2^2
}
\boldsymbol{\Delta}_{m,b}^{(t)}.
\end{equation}
This projection removes only the component opposing the current
aggregation direction. Since the safe reference in
\eqref{equ:tca_safe_ref} is non-conflicting with the current
aggregation direction, it does not reduce the required
alignment with the safe reference.

After assembling the encoder and decoder blocks, the final global
model is updated as
\begin{equation}
\label{equ:tca_final}
\mathbf{w}^{(t+1)}
=
\mathbf{w}^{(t+\frac{1}{2})}
-
\widetilde{\boldsymbol{\Delta}}_f^{(t)}.
\end{equation}
The overall procedure is summarized in Algorithm~\ref{alg:tca}.

\begin{algorithm}[t]
\caption{TCA Controlled FR}
\label{alg:tca}
\begin{algorithmic}[1]
\Require Global model $\mathbf{w}^{(t)}$;
aggregated model $\mathbf{w}^{(t+\frac{1}{2})}$;
raw FR model $\mathbf{w}_{\mathrm{raw}}^{(t+1)}$;
EMA states $\{\mathbf{h}_b^{(t-1)}\}$;
$a_{\mathrm{th}}$, $P_{\mathrm{act}}$, $\beta$, $c_{\mathrm{tar}}$

\State Compute $\boldsymbol{\Delta}_m^{(t)}$ and
$\boldsymbol{\Delta}_{f,\mathrm{raw}}^{(t)}$
using Eq.~\eqref{equ:tca_direction}

\For{$b\in\{\mathrm{enc},\mathrm{dec}\}$}
    \State Update $\mathbf{h}_b^{(t)}$ and
    $\mathbf{r}_b^{(t)}$ using Eq.~\eqref{equ:tca_ema}
\EndFor

\State Compute $a^{(t)}$ using Eq.~\eqref{equ:tca_cos}
and update the activation state

\If{TCA is inactive}
    \State $\mathbf{w}^{(t+1)}
    \leftarrow
    \mathbf{w}_{\mathrm{raw}}^{(t+1)}$
\Else
    \For{$b\in\{\mathrm{enc},\mathrm{dec}\}$}
        \State Select $\mathbf{r}_{b,\mathrm{safe}}^{(t)}$
        using Eq.~\eqref{equ:tca_safe_ref}

        \If{$
        \cos\left(
        \boldsymbol{\Delta}_{f,\mathrm{raw},b}^{(t)},
        \mathbf{r}_{b,\mathrm{safe}}^{(t)}
        \right)
        < c_{\mathrm{tar}}$}
            \State Align $\boldsymbol{\Delta}_{f,\mathrm{raw},b}^{(t)}$
            toward $\mathbf{r}_{b,\mathrm{safe}}^{(t)}$ to obtain $\boldsymbol{\Delta}_{f,b}^{(t)}$
        \Else
            \State $\overline{\boldsymbol{\Delta}}_{f,b}^{(t)}
            \leftarrow
            \boldsymbol{\Delta}_{f,\mathrm{raw},b}^{(t)}$
        \EndIf

        \State Apply Eq.~\eqref{equ:tca_safety} to obtain
        $\widetilde{\boldsymbol{\Delta}}_{f,b}^{(t)}$
    \EndFor

    \State Assemble $\widetilde{\boldsymbol{\Delta}}_f^{(t)}$
    from the encoder and decoder blocks

    \State $\mathbf{w}^{(t+1)}
    \leftarrow
    \mathbf{w}^{(t+\frac{1}{2})}
    -
    \widetilde{\boldsymbol{\Delta}}_f^{(t)}$
\EndIf

\State \Return $\mathbf{w}^{(t+1)}$
\end{algorithmic}
\end{algorithm}

\section{Numerical Experiments}
\subsection{Experimental Setup}
We evaluate ARC-FedSC on CIFAR-10 using PSNR as the image reconstruction metric. We set $K=50$ UEs, with each UE holding $|\mathcal{D}_k|=800$ training images, including 128 public samples for feature extraction. The CIFAR-10 training samples are randomly partitioned among UEs without label-based constraints, resulting in an approximately IID data setting. The server additionally maintains a balanced anchor set of 1000 unused training images for Role-Bandit evaluation. The JSCC feature dimension is 256 and the semantic channel SNR is fixed
at 20 dB. The mini-batch size is 16, and the numbers of local and server FR epochs are 3 and 5, respectively. The UE and server learning rates are initialized as $10^{-3}$ and $10^{-4}$ and are reduced by a factor of $0.8$ every 10 global rounds. The model update sparsification ratio is $S=0.4$, and all experiments run for 200 global rounds.

For the RAN uplink, we set $B=1$ MHz, $\rho=0.92$, and $T_{\max}=1.33$ s. The normalized role costs are set to $(\omega_m,\omega_o)=(4,1)$ with $C_{\max}=50$, consistent with the relative transmitted data sizes $S_m/N=0.4$ and $S_o/N=0.1$ for model updates and semantic features~\cite{huh2025feature}.
Accordingly, we consider $(K_m,K_o)\in\{(8,18),(9,14),(10,10),(11,6),(12,2)\}$, all satisfying $\omega_m K_m+\omega_o K_o=C_{max}$. Each role configuration remains fixed within an experiment, whereas the selected UEs and their assigned roles vary across rounds. Partial selection is allowed when the requested role quotas cannot be fully filled under the hard deadline. The main Role-Bandit parameters are 5 coverage rounds, a calibration interval of five rounds, a discount factor $\lambda=0.97$, an exploration factor $\xi_{\exp}=1$ and a staleness cap of $H=20$. For TCA, we set the activation threshold $a_{\mathrm{th}}=0.05$, activation patience $P_{\mathrm{act}}=2$, EMA factor $\beta=0.90$, and target cosine $c_{\mathrm{tar}}=0.05$. Formal experiments are conducted with random seeds $42$, $123$, and $2026$. 

\subsection{Baselines}

We compare ARC-FedSC with the following baselines under the
same JSCC and RAN settings.

\begin{itemize}
    \item \textbf{FedSFR}~\cite{huh2025feature}: FedSFR adapted to the same dynamic RAN
    environment and communication budget. UE role assignment is based
    only on instantaneous uplink channel conditions, and FR is directly
    performed at the server.

    \item \textbf{MFedMC}~\cite{yuan2026communication}: MFedMC adapted to the same two role setting under the dynamic RAN environment and communication budget. The original modality selection criteria are mapped to model and feature roles based on contribution, communication cost, and recency, while UE selection follows the lower local loss criterion.

    \item \textbf{FedAvg}~\cite{mcmahan2017communication}: A communication efficient FedAvg baseline
    using Top-$S$ sparsification and error feedback. Only model updates
    are transmitted, and 12 UEs are selected per round to ensure a
    comparable communication cost.
\end{itemize}

\subsection{Numerical Results}

\begin{figure}
    \centering
    \includegraphics[width=1\linewidth]{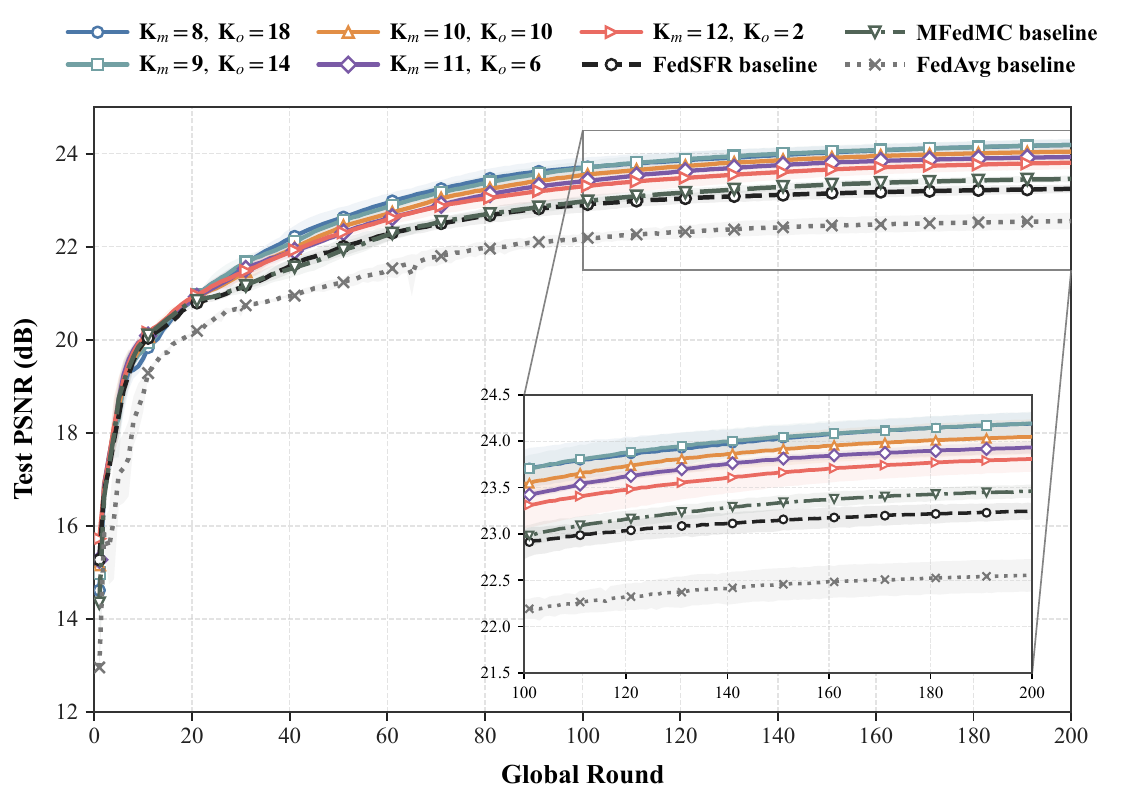}
    \caption{Test PSNR under different model-role and feature-role configurations of ARC-FedSC, comparing with baselines.}
    \label{fig:2}
\end{figure}

Fig.~\ref{fig:2} compares the PSNR performance across three random seeds. All five ARC-FedSC configurations outperform FedSFR, MFedMC, and FedAvg in later rounds. The $(8,18)$ and $(9,14)$ configurations achieve approximately $24.17$ dB last-20 PSNR, while further increasing $K_m$ does not improve performance under the fixed budget.

\begin{figure}
    \centering
    \includegraphics[width=1\linewidth]{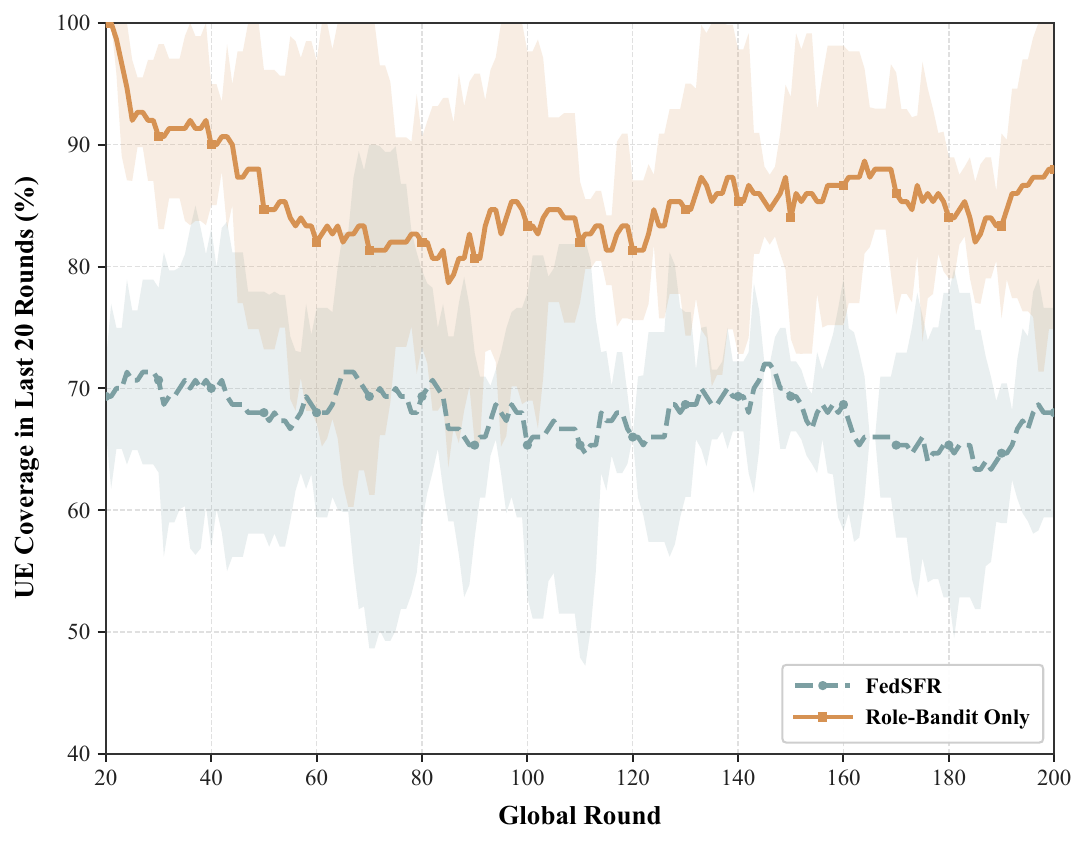}
    \caption{20-round rolling UE coverage. Solid/dashed lines denote the mean over three random seeds, and shaded regions indicate ±1 standard deviation.}
    \label{fig:3}
\end{figure}

\begin{figure}
    \centering
    \includegraphics[width=1\linewidth]{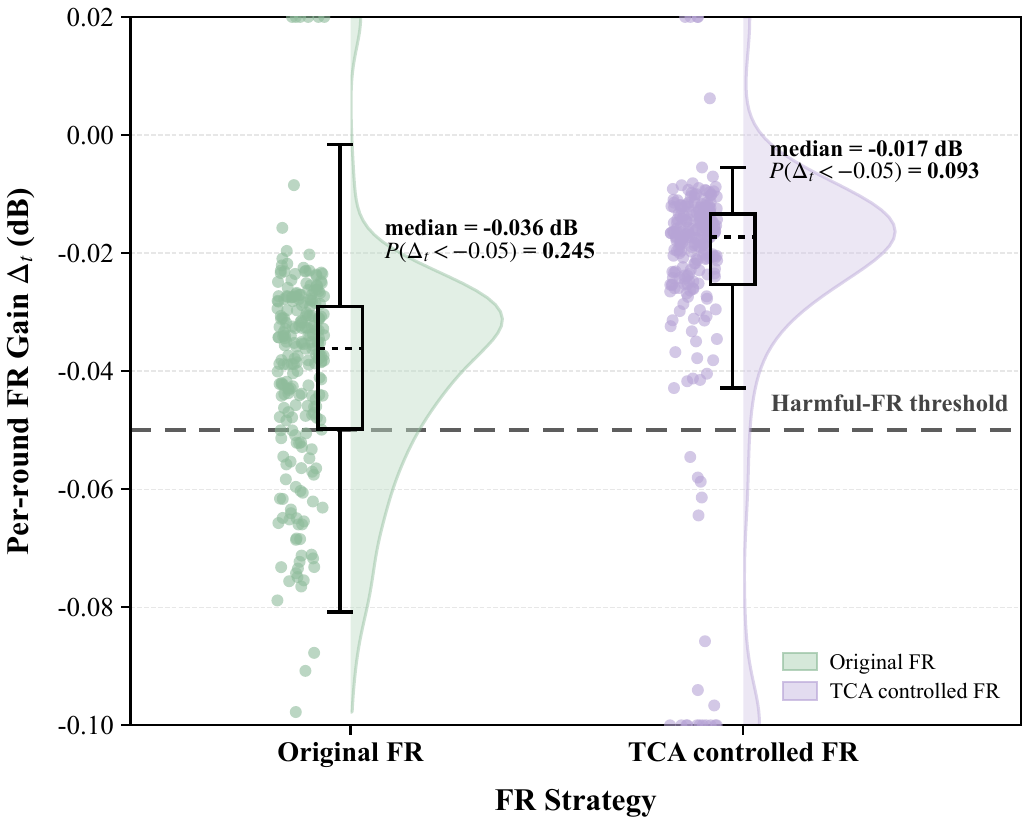}
    \caption{Distribution of per-round FR gain for the original FR and TCA controlled FR across training rounds. The dashed line at $\Delta_t=-0.05$ dB explicitly marks the threshold used to identify harmful FR events.}
    \label{fig:4}
\end{figure}

Fig.~\ref{fig:3} evaluates the effect of Role-Bandit on UE
participation. Compared with FedSFR, Role-Bandit maintains consistently
higher 20-round rolling UE coverage throughout training,
typically around $80$--$90\%$, while FedSFR remains around
$65$--$70\%$. This indicates that Role-Bandit promotes broader UE
participation and avoids repeatedly concentrating the training process
on a limited subset of UEs. Fig.~\ref{fig:4} evaluates the effect of TCA on FR at the server, where the FR gain in round $t$, denoted by $\Delta_t$, is
defined as the PSNR difference before and after FR. Since small negative gains close to zero may reflect minor performance fluctuations, we use $\Delta_t<-0.05$ dB to identify clearly harmful FR events. Compared with the original FR, TCA controlled FR shifts the gain distribution closer to zero and reduces harmful FR
updates. Specifically, the median gain
improves from $-0.036$ dB to $-0.017$ dB, while the probability of
$\Delta_t<-0.05$ dB decreases from $0.245$ to $0.093$, corresponding
to a $62.0\%$ reduction in harmful FR events. Meanwhile, the fraction of positive FR updates remains nearly unchanged, increasing only from $0.027$ to $0.030$. These results show that
TCA mainly improves FR by suppressing harmful updates.

We evaluate the server side LOO calibration overhead under \((K_m,K_o)=(10,10)\). Across 200 rounds and three random seeds, LOO computation takes an average of \(214.53\) s (\(1.07\) s per round), equivalent to \(3.67\%\) of the runtime of FedSFR in matched runs on the same hardware. Since LOO evaluation reuses received model updates and semantic features, it requires no additional uplink payload.

\subsection{Ablation Studies}

\textbf{Module Ablation.}
Table~\ref{tab:ablation} presents the module ablation results under
$(K_m,K_o)=(10,10)$, which is reported as a balanced configuration in FedSFR~\cite{huh2025feature}. Both Role-Bandit and TCA individually
improve FedSFR, increasing the last 20-round PSNR from $23.231$ dB
to $23.683$ dB and $23.736$ dB, respectively. Combining the two modules achieves the highest average PSNR of $24.030$ dB and requires only $98.7$ rounds to reach $23.5$ dB,
compared with $139.0$ rounds for Role-Bandit only and $118.7$ rounds for TCA only. Under the same communication budget, ARC-FedSC therefore achieves better reconstruction performance and reaches the target PSNR in fewer training rounds.

\begin{table}[t]
\centering
\caption{Ablation results under $(K_m,K_o)=(10,10)$ averaged over three random seeds.}
\label{tab:ablation}
\begin{tabular}{lcc}
\hline
Method & Last-20 PSNR (dB) & Round to 23.5 dB \\
\hline
FedAvg
& $22.540 \pm 0.176$
& $>200$ \\

FedSFR
& $23.231 \pm 0.096$
& $>200$ \\

Role-Bandit only
& $23.683 \pm 0.125$
& $139.0$ \\

TCA only
& $23.736 \pm 0.217$
& $118.7$ \\

\textbf{ARC-FedSC}
& $\mathbf{24.030 \pm 0.186}$
& $\mathbf{98.7}$ \\
\hline
\end{tabular}
\end{table}

\textbf{Role-Bandit Component Ablation.}
To examine the selection mechanism in more detail, we remove the exploration,
staleness, and communication terms individually while
retaining the historical reward term.
As shown in Table~\ref{tab:role_bandit_ablation}, removing any
of the three terms degrades the average reconstruction performance.
In particular, removing the communication term causes the
largest PSNR reduction, indicating the importance of accounting for communication conditions in role selection.
The exploration and staleness terms also improve performance by
maintaining selection diversity and preventing some UEs from being
ignored for a long time. In addition, the full Role-Bandit
achieves the highest average PSNR and the smallest variation across
the three random seeds.

\begin{table}[t]
\centering
\caption{Component ablation of Role-Bandit under
$(K_m,K_o)=(10,10)$ averaged over three random seeds.}
\label{tab:role_bandit_ablation}
\setlength{\tabcolsep}{6pt}
\begin{tabular}{lcc}
\hline
Variant & Last-20 PSNR (dB) & $\Delta$ PSNR (dB) \\
\hline
w/o Communication & $23.587 \pm 0.217$ & $-0.096$ \\
w/o Exploration   & $23.608 \pm 0.192$ & $-0.075$ \\
w/o Staleness     & $23.613 \pm 0.168$ & $-0.070$ \\
\textbf{Full Role-Bandit}
                  & $\mathbf{23.683 \pm 0.125}$ & -- \\
\hline
\end{tabular}
\end{table}

\textbf{Parameter Sensitivity.}
After validating the module and component designs, we examine the
sensitivity to the main parameters of Role-Bandit and TCA.
For Role-Bandit, the discount factor $\lambda$ controls the
contribution of historical observations, while the exploration
factor $\xi_{\exp}$ scales the default exploration coefficients as
$\beta_m=0.025\xi_{\exp}$ and
$\beta_o=0.035\xi_{\exp}$.
As shown in the upper part of Table~\ref{tab:parameter_sensitivity},
$\lambda=0.97$ gives the highest PSNR among the tested values, suggesting
a balance between historical information and recent observations. The PSNR varies only slightly when
$\xi_{\exp}$ changes from $0.5$ to $2.0$,
demonstrating the robustness of the Role-Bandit mechanism to the
exploration strength. For TCA, we study the EMA factor $\beta$ used to construct the
temporal reference direction.
As shown in the lower part of Table~\ref{tab:parameter_sensitivity}, setting $\beta=0$,
which removes temporal smoothing, gives the lowest average PSNR.
All tested nonzero values improve the average reconstruction performance,
with $\beta=0.9$ achieving the highest last-20-round PSNR.
The comparable performance for $\beta=0.5$ and $\beta=0.9$
also indicates that TCA is relatively insensitive within this range.
When $\beta$ increases to $0.99$, the PSNR
decreases slightly, suggesting that excessive historical smoothing may reduce
adaptation to recent model updates.

\begin{table}[t]
\centering
\caption{Parameter sensitivity of Role-Bandit UE selection and TCA FR under
$(K_m,K_o)=(10,10)$. Default parameter
settings are shown in bold.}
\label{tab:parameter_sensitivity}
\setlength{\tabcolsep}{7pt}
\renewcommand{\arraystretch}{1.1}

\begin{tabular}{ccc}
\hline
\textbf{Role-Bandit UE selection} & \textbf{Parameter Value} & \textbf{Last-20 PSNR (dB)} \\
\hline
\multirow[c]{5}{*}{Discount factor $\lambda$}
& 0.90 & 23.636 \\
& 0.95 & 23.637 \\
& \textbf{0.97} & \textbf{23.683} \\
& 0.99 & 23.657 \\
& 1.00 & 23.630 \\
\hline
\multirow[c]{5}{*}{Exploration
factor $\xi_{\exp}$}
& 0.50 & 23.664 \\
& 0.75 & 23.655 \\
& \textbf{1.00} & \textbf{23.683} \\
& 1.50 & 23.657 \\
& 2.00 & 23.670 \\
\hline
\textbf{TCA FR} & \textbf{Parameter Value} & \textbf{Last-20 PSNR (dB)} \\
\hline
\multirow[c]{4}{*}{EMA factor $\beta$}
& 0.00 & 23.696 \\
& 0.50 & 23.699 \\
& \textbf{0.90} & \textbf{23.736} \\
& 0.99 & 23.708 \\
\hline
\end{tabular}
\end{table}

Taken together, these results support the design of both Role-Bandit
and TCA and show stable performance over the tested parameter ranges.

\section{Conclusion}
In this paper, we proposed ARC-FedSC to coordinate UE role assignment and server side FR under constrained RAN resources. In our proposed framework, role-Bandit combines role-specific historical contributions with current uplink conditions, while TCA regulates FR using recent aggregation directions. Experiments demonstrated improves reconstruction quality, UE coverage, convergence speed, and FR reliability over the baselines. Ablation and sensitivity studies further validated the effectiveness and robustness of both components.

\bibliographystyle{IEEEtran}
\bibliography{ref}

\begin{thebibliography}{10}
\providecommand{\url}[1]{#1}
\csname url@rmstyle\endcsname
\providecommand{\newblock}{\relax}
\providecommand{\bibinfo}[2]{#2}
\providecommand\BIBentrySTDinterwordspacing{\spaceskip=0pt\relax}
\providecommand\BIBentryALTinterwordstretchfactor{4}
\providecommand\BIBentryALTinterwordspacing{\spaceskip=\fontdimen2\font plus
\BIBentryALTinterwordstretchfactor\fontdimen3\font minus \fontdimen4\font\relax}
\providecommand\BIBforeignlanguage[2]{{%
\expandafter\ifx\csname l@#1\endcsname\relax
\typeout{** WARNING: IEEEtran.bst: No hyphenation pattern has been}%
\typeout{** loaded for the language `#1'. Using the pattern for}%
\typeout{** the default language instead.}%
\else
\language=\csname l@#1\endcsname
\fi
#2}}

\bibitem{xie2021deep}
H.~Xie, Z.~Qin, G.~Y. Li, and B.-H. Juang, ``Deep learning enabled semantic communication systems,'' \emph{IEEE Trans. Signal Process.}, vol.~69, pp. 2663--2675, 2021.

\bibitem{bourtsoulatze2019deep}
E.~Bourtsoulatze, D.~B. Kurka, and D.~G{\"u}nd{\"u}z, ``Deep joint source-channel coding for wireless image transmission,'' \emph{IEEE Trans. Cogn. Commun. Netw.}, vol.~5, no.~3, pp. 567--579, 2019.

\bibitem{dai2022nonlinear}
J.~Dai, S.~Wang, K.~Tan, Z.~Si, X.~Qin, K.~Niu, and P.~Zhang, ``Nonlinear transform source-channel coding for semantic communications,'' \emph{IEEE J. Sel. Areas Commun.}, vol.~40, no.~8, pp. 2300--2316, 2022.

\bibitem{huh2025feature}
Y.~Huh, B.~Kim, and W.~Choi, ``Feature reconstruction aided federated learning for image semantic communication,'' in \emph{Proc. IEEE Global Commun. Conf. (GLOBECOM)}.\hskip 1em plus 0.5em minus 0.4em\relax IEEE, 2025, pp. 4669--4674.

\bibitem{mcmahan2017communication}
B.~McMahan, E.~Moore, D.~Ramage, S.~Hampson, and B.~A. y~Arcas, ``Communication-efficient learning of deep networks from decentralized data,'' in \emph{Proc. 20th Int. Conf. Artif. Intell. Statist. (AISTATS)}.\hskip 1em plus 0.5em minus 0.4em\relax Pmlr, 2017, pp. 1273--1282.

\bibitem{Qi2025}
H.~Qi, J.~Luo, Z.~Xu, Q.~Li, J.~Yin, and J.~Wu, ``Energy efficient power control for over-the-air federated learning in satellite communications,'' \emph{IEEE Commun. Lett.}, vol.~29, no.~7, pp. 1530--1534, 2025.

\bibitem{sun2024federated}
H.~Sun, H.~Tian, W.~Ni, and J.~Zheng, ``Federated learning-based cooperative model training for task-oriented semantic communication,'' in \emph{Proc. IEEE INFOCOM Workshops}, 2024, pp. 1--6.

\bibitem{nguyen2024efficient}
L.~X. Nguyen, H.~Q. Le, Y.~L. Tun, P.~S. Aung, Y.~K. Tun, Z.~Han, and C.~S. Hong, ``An efficient federated learning framework for training semantic communication systems,'' \emph{IEEE Trans. Veh. Technol.}, vol.~73, no.~10, pp. 15\,872--15\,877, 2024.

\bibitem{yan2024fssc}
Y.~Yan, X.~Zhang, L.~Li, W.~Lin, R.~Li, W.~Cheng, and Z.~Han, ``Fssc: Federated learning of transformer neural networks for semantic image communication,'' in \emph{Proc. IEEE Global Commun. Conf. (GLOBECOM)}.\hskip 1em plus 0.5em minus 0.4em\relax IEEE, 2024, pp. 1659--1664.

\bibitem{xu2024federated}
J.~Xu, H.~Yao, R.~Zhang, T.~Mai, S.~Huang, and S.~Guo, ``Federated learning powered semantic communication for uav swarm cooperation,'' \emph{IEEE Wireless Commun.}, vol.~31, no.~4, pp. 140--146, 2024.

\bibitem{zheng2024mobility}
G.~Zheng, Q.~Ni, K.~Navaie, H.~Pervaiz, G.~Min, A.~Kaushik, and C.~Zarakovitis, ``Mobility-aware split-federated with transfer learning for vehicular semantic communication networks,'' \emph{IEEE Internet Things J.}, vol.~11, no.~10, pp. 17\,237--17\,248, 2024.

\bibitem{yu2026scusfl}
L.~Yu and Z.~Chang, ``Semantic communication-enhanced u-shaped split federated learning with adaptive compression for vehicular networks,'' \emph{IEEE Trans. Wireless Commun.}, vol.~25, pp. 11\,609--11\,623, 2026.

\bibitem{amiri2020federated}
M.~M. Amiri and D.~G{\"u}nd{\"u}z, ``Federated learning over wireless fading channels,'' \emph{IEEE Trans. Wireless Commun.}, vol.~19, no.~5, pp. 3546--3557, 2020.

\bibitem{nishio2019client}
T.~Nishio and R.~Yonetani, ``Client selection for federated learning with heterogeneous resources in mobile edge,'' in \emph{Proc. IEEE Int. Conf. Commun. (ICC)}, 2019, pp. 1--7.

\bibitem{xu2020client}
J.~Xu and H.~Wang, ``Client selection and bandwidth allocation in wireless federated learning networks: A long-term perspective,'' \emph{IEEE Trans. Wireless Commun.}, vol.~20, no.~2, pp. 1188--1200, 2021.

\bibitem{xu2023federated}
Y.~Xu, Z.~Jiang, H.~Xu, Z.~Wang, C.~Qian, and C.~Qiao, ``Federated learning with client selection and gradient compression in heterogeneous edge systems,'' \emph{IEEE Trans. Mobile Comput.}, vol.~23, no.~5, pp. 5446--5461, 2024.

\bibitem{yuan2026communication}
L.~Yuan, D.-J. Han, S.~Wang, D.~Upadhyay, and C.~G. Brinton, ``Communication-efficient multimodal federated learning: Joint modality and client selection,'' \emph{IEEE Trans. Mobile Comput.}, vol.~25, no.~8, pp. 12\,780--12\,796, 2026.

\bibitem{sun2025hybrid}
H.~Sun, W.~Ni, H.~Tian, J.~Zheng, G.~Nie, and D.~Niyato, ``A hybrid federated learning framework for task-oriented semantic communication,'' \emph{IEEE Internet Things J.}, vol.~12, no.~13, pp. 23\,444--23\,461, 2025.

\bibitem{xu2026flsc}
S.~Xu, Y.~Qi, F.~Qi, S.~Guo, and Z.~Zhao, ``Flsc-ci: Federated learning and semantic communication empowered multimodal terminal collaborative inferencing framework for iot businesses,'' \emph{IEEE Trans. Netw. Sci. Eng.}, 2026.

\bibitem{10599557}
X.~Xu, Y.~Xu, H.~Dou, M.~Chen, and L.~Wang, ``Federated kd-assisted image semantic communication in iot edge learning,'' \emph{IEEE Internet Things J.}, vol.~11, no.~21, pp. 34\,215--34\,228, 2024.

\bibitem{lai2021oort}
F.~Lai, X.~Zhu, H.~V. Madhyastha, and M.~Chowdhury, ``Oort: Efficient federated learning via guided participant selection,'' in \emph{Proc. 15th USENIX Symp. Oper. Syst. Des. Implement. (OSDI)}, 2021, pp. 19--35.

\bibitem{xia2020multi}
W.~Xia, T.~Q. Quek, K.~Guo, W.~Wen, H.~H. Yang, and H.~Zhu, ``Multi-armed bandit-based client scheduling for federated learning,'' \emph{IEEE Trans. Wireless Commun.}, vol.~19, no.~11, pp. 7108--7123, 2020.

\bibitem{auer2002finite}
P.~Auer, N.~Cesa-Bianchi, and P.~Fischer, ``Finite-time analysis of the multiarmed bandit problem,'' \emph{Mach. Learn.}, vol.~47, no.~2, pp. 235--256, 2002.

\end{thebibliography}

\vfill


\end{document}